\documentclass[10pt,twocolumn,letterpaper]{article}

\usepackage{iccv}
\usepackage{times}
\usepackage{epsfig}
\usepackage{graphicx}
\usepackage{amsmath}
\usepackage{amssymb}
\usepackage{multirow}
\usepackage{algorithm, algorithmic}
\usepackage[accsupp]{axessibility}

\usepackage[pagebackref=true,breaklinks=true,letterpaper=true,colorlinks,bookmarks=false]{hyperref}

\iccvfinalcopy 

\def\iccvPaperID{5515} 

\begin{document}

\title{Temporal Knowledge Consistency for Unsupervised Visual Representation Learning}

\author{Weixin Feng\textsuperscript{1}\thanks{Equal Contribution.}\ \ \ \  Yuanjiang Wang\textsuperscript{2}\footnotemark[1]\ \    \thanks{Corresponding author.}  \ \ \  Lihua Ma\textsuperscript{2} \ \ \ Ye Yuan\textsuperscript{2} \ \ Chi Zhang\textsuperscript{2}\\
Beijing University of Posts and Telecommunications \textsuperscript{1}\ \ \ \ Megvii Technology\textsuperscript{2}\\
{\tt\small fengweixin@bupt.edu.cn, wangyuanjiang@megvii.com} \\
{\tt\small  \{malihua, yuanye, zhangchi\}@megvii.com}
\vspace{-0.3em}\\
}

\maketitle
{\let\thefootnote\relax\footnote{{This paper is supported by the National Key R\&D Plan of the Ministry of Science and Technology (Project No.2020AAA0104400).}}}

\ificcvfinal\thispagestyle{empty}\fi



\begin{abstract}
The instance discrimination paradigm has become dominant in unsupervised learning. It always adopts a teacher-student framework, in which the teacher provides embedded knowledge as a supervision signal for the student. The student learns meaningful representations by enforcing instance spatial consistency with the views from the teacher. However, the outputs of the teacher can vary dramatically on the same instance during different training stages, introducing unexpected noise and leading to catastrophic forgetting caused by inconsistent objectives. In this paper, we first integrate instance temporal consistency into current instance discrimination paradigms, and propose a novel and strong algorithm named Temporal Knowledge Consistency (TKC). Specifically, our TKC dynamically ensembles the knowledge of temporal teachers and adaptively selects useful information according to its importance to learning instance temporal consistency. Experimental result shows that TKC can learn better visual representations on both ResNet and AlexNet on linear evaluation protocol while transfer well to downstream tasks. All experiments suggest the good effectiveness and generalization of our method. Code will be made available.
\end{abstract}




\section{Introduction}
\begin{figure}[ht]
\begin{center}
\includegraphics[width=0.5\textwidth]{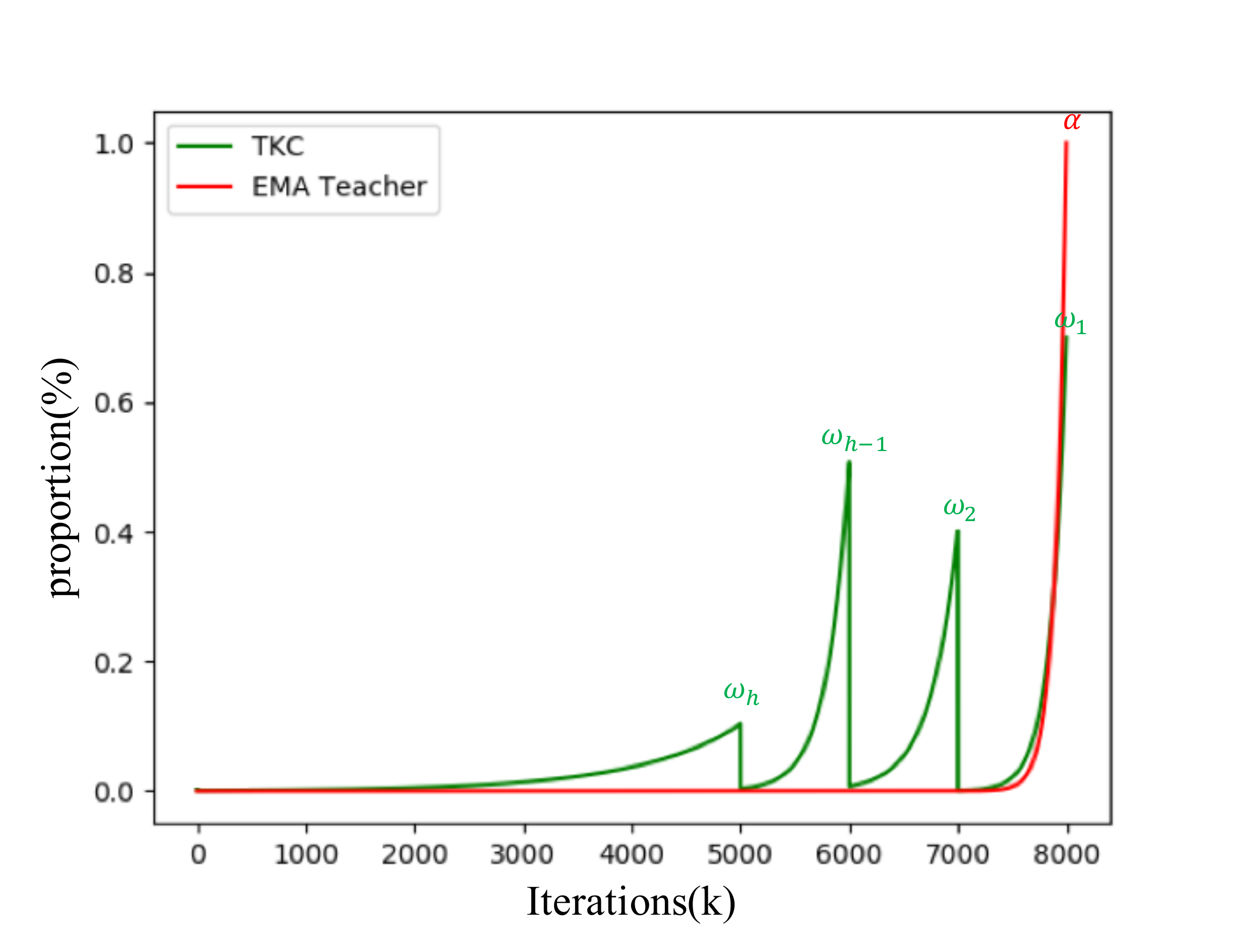}
\caption{Mainstream unsupervised methods adopt the teacher-student framework, where the teacher is an EMA ensemble of previous student encoders. This figure illustrates the proportion of previous students in the teacher with respect to training steps. The red curve shows that the EMA teacher ensembles the previous encoders by a predesigned factor $\alpha$, where only alomst encoders in the very close steps are ensembled. Our TKC (the green curve) reuses the early models and adaptively learns the importance $\omega $ for each of them, thus leads to temporal consistent representations.}
\vspace{-3mm}
\label{fig:mot}
\end{center}
\end{figure}

The rise of Deep Convolutional Neural Networks (DCNN) \cite{resnet, alexnet, vgg} has led to significant success in computer vision benchmarks \cite{imagenet, voc, coco}. The excellent performance of supervised DCNN always relies on a large quantity of manually labeled data, which is costly to collect \cite{RotNet, NPID}. Unsupervised representation learning has been attracted more and more interest, for it can learn a good representation without human annotations. These methods are generally to manually design a pretext task to learn representations, such as image in-painting \cite{paint}, colorization \cite{color1,color2,color3,color4}, rotate predicting \cite{RotNet,Rot2,Rot3} and clustering \cite{DC, LA, SWAV}. All these pretext tasks are based on specific domain knowledge, which has poor generation on various downstream tasks. Recently, instance discrimination \cite{NPID, MOCO, SimCLR, BYOL, PIRL} paradigm has led to remarkable progress in unsupervised representation learning and even surpasses the supervised pre-training on extensive downstream tasks \cite{PIRL, MOCO}. 

The instance discrimination paradigm treats each sample itself as its own category and trains the CNN to separate all the different samples from each other. The current paradigm can be formulated as a teacher-student framework enforcing the instance spatial consistency of two networks, which are the student network and the EMA teacher network \cite{MOCO_v2, BYOL, SimCLR}. The instance spatial consistency constrains the similarity of different spatial views from the same instance, and its ultimate goal is to learn instance-discriminative and spatial-invariant representations. One of the key points in these instance discrimination works is the EMA teacher. For instance, MoCo \cite{MOCO} uses the EMA teacher to output consistent negative samples for the student; BYOL \cite{BYOL} trains a student to mimic the representations from the EMA teacher; SimCLR \cite{SimCLR} maintains a real-time EMA teacher of the student.

However, we argue that the current EMA teacher is sub-optimal as illustrated in Fig. \ref{fig:mot}: (1) the EMA teacher only ensembles the rare knowledge of recent encoders by a handcraft proportion, which means that it only concentrates on instance spatial consistency while the instance temporal consistency is ignored. As a consequence, the outputs of the same sample can vary dramatically among different training stages, which can introduce unexpected noise and finally lead to catastrophic forgetting \cite{forget, TCSSL}. (2) The EMA manner can’t leverage the importance of different encoders. It assumes that the outputs of later models are largely more important than the earlier ones, despite that the benefits of previous epochs have been observed in previous works \cite{PI, TCSSL}.


In this paper, we integrate instance temporal consistency into the instance discrimination paradigm and propose a novel and strong algorithm, namely Temporal Knowledge Consistency(TKC), which contains the \textit{temporal teacher} and the \textit{knowledge transformer}. Specifically, \textit{temporal teacher} supplies instance temporal consistency via introducing the temporal knowledge from previous models. And the \textit{knowledge transformer} dynamically learns the importance of different temporal teachers, then adaptively ensembles the useful information according to their importance, to generate instance temporal consistency objective. In addition, we provide a computation-economical implementation, which can provide temporal knowledge without preserving multiple previous models.

Our experimental results on different tasks and benchmarks have demonstrated that TKC can learn a better visual representation with excellent transferability and scalability. Concretely, we achieve state-of-the-art performance on ResNet and AlexNet backbones on linear evaluation protocol. Moreover, we evaluate representations learned by TKC on many downstream tasks and architectures. All results suggest the effectiveness of TKC. Overall, the main contributions in this work include:
\begin{itemize}
\item We are the first to integrate instance temporal consistency into the current EMA teacher in the instance discrimination paradigm.
\item We propose a novel and strong algorithm, named Temporal Knowledge Consistency (TKC), which can dynamically ensemble the knowledge from different temporal teachers. 
\item Extensive experiments are conducted on several benchmarks and architectures, which shows the superior performance on mainstream benchmarks and the scalability of TKC.
\end{itemize}

\section{Related Works}

\noindent\textbf{Unsupervised Pretext Tasks.} Unsupervised representation learning aims to learn meaningful representations from large amounts of data samples via constructing a wide range of pretext tasks without human labels. These pretext tasks usually vary in different forms. Among them, one family of these typical pretext works are generative-based which rely on auto-encoder \cite{ae} or GAN \cite{gan1,gan2}, such as colorization \cite{color1,color2,color3,color4} and image in-painting \cite{paint}. And the others are discriminative-based, like predicting rotation or augmentation \cite{RotNet,Rot2,Rot3} of the image, solving jigsaw puzzles \cite{jigsaw}, locating relative patch \cite{patch1,patch2}, ordering video frames \cite{video1,video2,video3}, matching corresponding audio \cite{audio1,audio2,audio3,audio4}, and clustering \cite{DC, DC2, LA, ODC, SELA, cluster1, cluster2, ClusterFit, SCAN}. All the pretext methods are based on specific domain knowledge, fail to generalize to different downstream tasks. Recent progress in unsupervised representation learning mainly benefits from instance discrimination and attracts widespread attention from researchers.

\noindent{\textbf{Instance Discrimination.}} Instance discrimination methods \cite{NPID, MOCO, MOCO_v2, SimCLR, BYOL, CPC1, Rot3, PIRL} have dominated the unsupervised learning field in the few years, which treat each sample itself as its own category and train the CNN to separate all the different samples from each other. This paradigm commonly includes a teacher model to provide a supervised signal, and a student model to learn the embedded knowledge from the former. Wu \etal \cite{NPID} is the first to propose instance discrimination in unsupervised learning, which regards the student model in the last epoch as the teacher model. It learns meaningful representations by means of the classic InfoNCE loss \cite{NCE, CPC1} and the target generated by the teacher. MoCo \cite{MOCO, MOCO_v2} takes the EMA ensemble of the student as the teacher model to provide consistent and robust objectives, and brings a breakthrough by solving the knowledge out-of-date problem with the help of the EMA teacher. It also maintains a queue of negative samples and keeps them fresh. SimCLR \cite{SimCLR} builds symmetrical architecture between the student and the teacher, while uses stronger data augmentation to enforce network to learn instance spatial consistency. BYOL \cite{BYOL} also implements the teacher with the EMA ensemble of the student, and makes use of L2 loss to pull the embedding features of positives pairs while removing explicitly negative samples. Our TKC explicitly integrates instance temporal consistency into the instance discrimination paradigm, making the targets generated by the teacher more accurate and stable.


\noindent{\textbf{Temporal Knowledge.}} Temporal knowledge is widely used in both semi-supervised learning and optimization. In the field of semi-supervised learning, plenty of the proposed works adopt the EMA ensemble to take advantage of the knowledge of the previous training stage, to learn the time-consistent representations. Temporal Ensemble \cite{PI} ensembles the output of different epochs to yield better predictions. Mean teacher \cite{MT} instead ensembles the previous student as a teacher to prevent incorrect target and outweighs the cost of misclassification. Tian \etal \cite{TCSSL} points out the catastrophic forgetting problems in semi-supervised, and solves it by measuring the time consistency of samples and filtering the inconsistent ones. In the field of optimization, temporal knowledge is integrated by different advanced optimization strategies during training. SGD only uses the gradient computed by a mini-batch to back-propagate, which is noisy and inaccurate. Momentum \cite{momentum} and NAG \cite{nag} instead use the gradient in a mini-batch by the momentum of the gradient to accelerate the convergence of model training and suppress shocks. Adam \cite{adam} is another momentum updating strategy which further introduces the second momentum to leverage different channels. All these works use temporal knowledge to reduce noise and accelerate convergence.

\begin{figure*}[ht]
\centering
\includegraphics[width=0.9\textwidth]{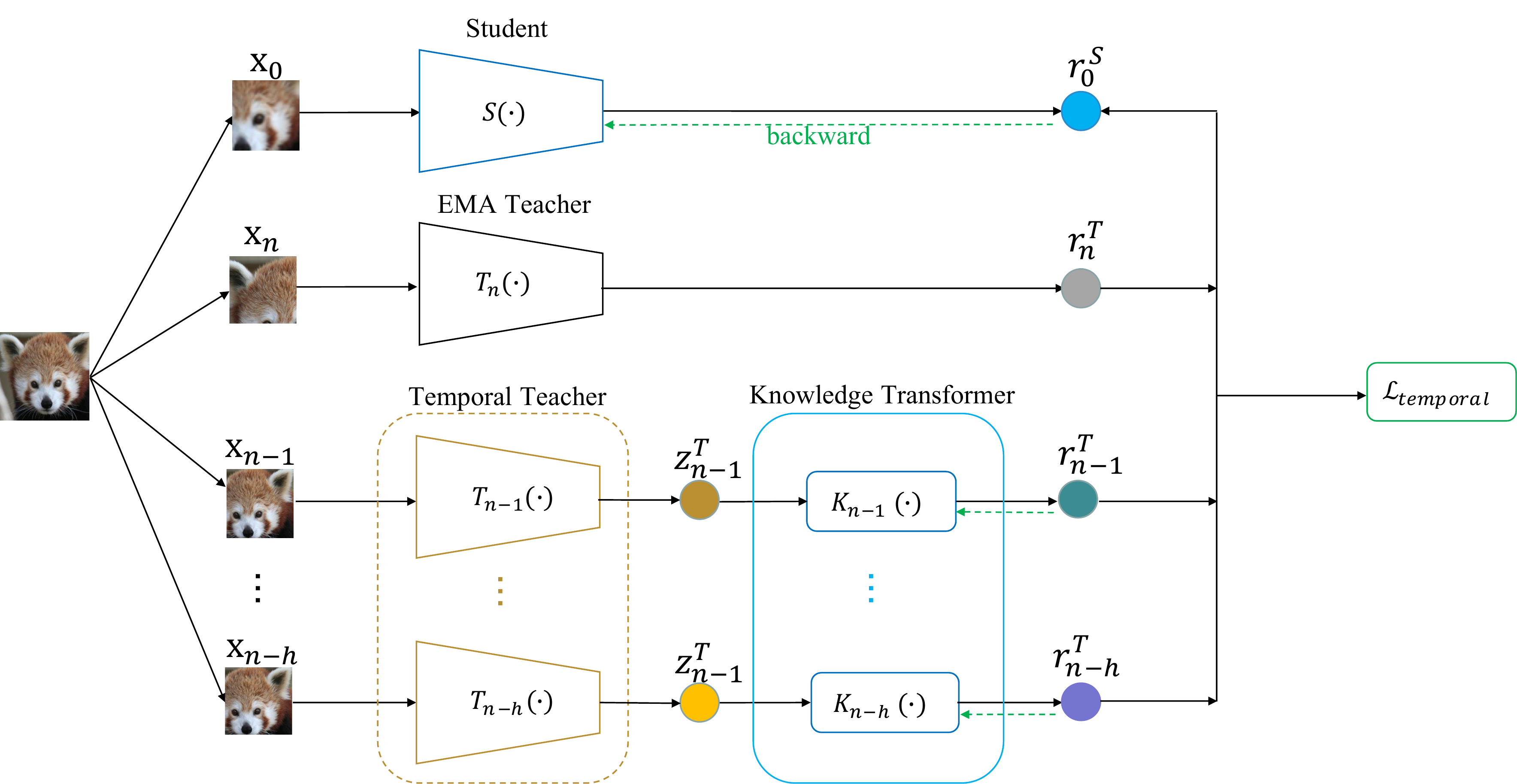}
\vspace{+4mm}
\caption{\textbf{The overall framework of our TKC.} For each training image $x $, TKC generate a target from the EMA teacher, and $h $ targets from the \textit{temporal teachers}. \textit{Temporal teachers} is a set of encoders from previous training stages. The \textit{knowledge transformer} is appended behind the \textit{temporal teacher} to dynamically leverage their importance. Every teacher in TKC frameworks can provide a supervised signal, which then feeds into the \textit{temporal loss} and backward to update the student and the \textit{knowledge transformer}. The green dotted line means backpropagate.}
\label{fig:pipeline2}
\end{figure*}

\section{Method}\label{method}

In this section, we first point out the limitation of the current EMA teacher in Sec \ref{method:EMATeacher}. Secondly, we propose \textit{temporal teacher} to improve it in Sec \ref{method:ST}. Thirdly, we introduce the \textit{knowledge transformer} to dynamically leverage the importance of different models in Sec \ref{method:KT}. Then we propose a \textit{temporal loss} to learn instance temporal consistency in Sec \ref{method:loss}.  At last, we describe our overall framework and the algorithm in Sec \ref{method:Integral}.

\subsection{Limitation of EMA teacher}\label{method:EMATeacher}

Instance discrimination paradigm always involves two encoders, the teacher encoder $T $ and the student encoder $S $. For a training sample $x $, the augmentation from augmentation distribution $\mathcal{T} $ is applied twice to obtain two augmented sample $x_{0} $, $x_{n} $. The teacher output $r_{n}^{T} = T(x_{n}) $ as the target to provide instance spatial knowledge. The student network takes the other sample $x_{0} $ then outputs $r_{0}^{S} = T(x_{0}) $, and learn knowledge by constrainting its similarity with $r_{n}^{T} $. In this teacher-student framework, the teacher encoder has the same architecture with the student, and its parameters are updated by an exponential moving average (EMA) of the models:
\begin{equation}\label{eq:EMA}
T^{n+1} = \alpha T^{n} + (1-\alpha )S^{n}
\end{equation}
where $n $ is the training step, $\alpha $ is to control the updating speed of the teacher. We name the teacher as EMA teacher. In current training step $n+1 $, the teacher is ensembled by the last teacher $T^{n} $ with ratio $\alpha $, and the last student $S^{n} $ with ratio $1-\alpha $. The last teacher $T^{n} $ is also an ensemble of previous students. In order to explore the temporal knowledge in EMA teacher, we expand $T^{n} $ in Eq \ref{eq:EMA} as following: 
\begin{equation}\label{eq:EMA_sp}
\begin{aligned}
T^{n+1} &= (1-\alpha) \cdot \sum_{m=0}^{n-1}(\alpha^{m}S^{n-m}) + \alpha^{n}\cdot T_{1}\\
&\approx (1-\alpha) \cdot[S^{n} +\alpha S^{n\!-\!1}+...+\alpha^{n} S^{0}]
\end{aligned}
\end{equation}
where $S^{m} $ means the student model at step $m $, $\alpha $ is the updating factor. In Eq \ref{eq:EMA_sp}, we can find out that current teacher $T^{n+1} $ is an ensemble of a sequence of student $S $ from step 0 to step $n $. 

However, we note that the EMA teacher can only preserve the knowledge from the latest encoders. On the one hand, as $m $ goes to infinity $m \to \infty$, the weight of student $S^{n-m} $ approaches 0, for $\alpha $ is lower than 1. When training MoCo \cite{MOCO} on ImageNet, only student models within an epoch can provide the knowledge, as illustrated in Fig. \ref{fig:mot}. This knowledge from only near steps is insufficient, which can cause the dramatically changes among different training stages and prevent the student to learn instance temporal consistency.

On the other hand, the strategy of EMA is also too simple. It assumes that the importance of earlier models is decreased exponentially with time, even though the earlier models can provide useful information to mitigate the catastrophic forgetting. In a summary, these two flaws prevent the instance discrimination paradigm from making full use of temporal knowledge and learning instance temporal consistency.

\subsection{Temporal Teachers}\label{method:ST}

EMA teacher in Eq.\ref{eq:EMA_sp} only attaches importance to recent models. However, the output of these models is smooth and similar due to the low learning rate and momentum optimizer. As a consequence, they fail to supply instance temporal consistency to lighten the dramatic changes of the models, which can easily lead to training failure and catastrophic forgetting. We claim that jointly utilizing the knowledge from previous models can provide a more consistent and robust target. To achieve that, we propose to take out the previous models, which have few proportions in the EMA teacher, to build our \textit{\textbf{temporal teacher}}. Then we make full use of them to alleviate catastrophic forgetting and learn instance temporal consistency in the instance discrimination paradigm.

We explicitly preserve a group of previous teachers as \textit{temporal teacher} to reuse the knowledge from previous encoders. To formulate our proposal, we use $T_{n} $ to denote current EMA teacher, and $\{T_{n-1}, T_{n-2},...\} $ to represent the temporal teachers. The lower subscript means earlier. Each of these teachers is saved for each $s $ training step, including the knowledge mainly from this training step. Note that the subscript $n $ means different with the superscript in Eq.\ref{eq:EMA}. The distance between $T_{j-1} $ and $T_{j-2} $ is $s $ training steps.

The teacher far away from now is too out-of-date, whose knowledge can be inconsistent and noisy for the current teacher. Hence we only preserve adjacent teachers as \textit{temporal teacher}, while throwing away the previous one. We use $h $ to represent the number of temporal teachers, and denote the \textit{temporal teachers} by $\{T_{n-1}, T_{n-2},..., T_{n-h} \} $.

We illustrate the \textit{temporal teacher} in the brown dotted box in Fig.\ref{fig:pipeline2}. For a sample $x $ in the training set, we apply $h $ times data augmentations from the augmentation distribution $\mathcal{T} $, to obtain $ x_{j}, j\in [n-h, n-1] $. The \textit{temporal teachers} are set stop-gradient, and take the augmented views as input to yield representations $z_{j}^{T}=T_{j}(x_{j}), j \in [n-h, n-1]$ as the temporal predicting target. The subscript $j $ of $z_{j}^{T} $ indicate that the target is corresponding to the teacher $T_{j} $.

In the implementation, we propose a more efficient way to achieve \textit{temporal teacher}. Instead of getting the target from previous teachers, we preserve the representations of all the training data in the previous $h $ stages in a memory named \textit{\textbf{history bank}}. For each training sample $x $, we can get $z_{j}^{T} $ from the \textit{history bank} instead of from the teacher $T_{j} $. \textit{History bank} is an approximate implementation of the \textit{temporal teachers}, for both of them can provide temporal knowledge. In this way, the computational cost is largely reduced and the additional GPU memory allocation is negligible. The detail can be seen in the supplementary material.

\subsection{Knowledge Transformer} \label{method:KT}

In EMA teacher, the weights of different ensemble models are decreased over time exponentially. However, the importance of different models may not be in line with the EMA rule. In this section, we propose to dynamically predict the importance of the \textit{temporal teachers}'s knowledge by \textit{\textbf{knowledge transformer}}. 

The \textit{knowledge transformer} is illustrated in the blue box in Fig.\ref{fig:pipeline2}. It takes the rough target $z_{j}^{T} ,j \in [n-1, n-h]$ from the teacher $T_{j} $ as input, and then transfer the knowledge of them to leverage their importance. The formulation is as follows:
\begin{equation} \label{eq:aw}
r_{j}^{T} = K_{j}(z_{j}^{T})
\end{equation}
where $r_{j}^{T} $ denotes the target after leveraging the importance, which has thrown the harmful information from it and only preserves temporal consistent knowledge. This strategy can adaptively learn and adjust the importance of the \textit{temporal teachers} in the early or later encoders, which is better than coupled it to the handcraft proportion in the EMA teacher. In the implementation, we use an MLP with one hidden layer to transfer the knowledge for each temporal teacher. During training, the \textit{knowledge transformer} is training simultaneously with the student. 

\subsection{Temporal Loss} \label{method:loss}

\begin{algorithm}[hbt]
\renewcommand{\algorithmicrequire}{\textbf{input:}}
\renewcommand{\algorithmicensure}{\textbf{hyperparameters:}}
\caption{Temporal knowledge consistency}\label{algorithm}
\begin{algorithmic}[1]
\REQUIRE $S(\cdot)$, $T_{n}(\cdot)$, $K(\cdot)$
\ENSURE $\alpha $, $h $, $s $
\FOR{each sample $x $}
\STATE draw h+2 augmentations
\STATE \textit{$\#$ the original models}
\STATE $r_{0}^{S} = S(x_{0}) $
\STATE $r_{n}^{T} = T_{n}(x_{n}) $
\FORALL{$j\in\{n-1,n-h\}$}
\STATE \textit{$\#$ temporal teacher}
\STATE $z_{j}^{T} = T_{j}(x_{j})$
\STATE \textit{$\#$ knowledge transformer}
\STATE $r_{j}^{T} = K_{j}(z_{j}^{T})$
\ENDFOR
\STATE \textit{$\#$ temporal loss}
\STATE compute the loss in Eq. \ref{eq:tem}
\STATE backward to update $S $ and $K $
\STATE update the $T_{n} $ by Eq. \ref{eq:EMA}
\ENDFOR
\STATE \textbf{return} $S(\cdot)$
\end{algorithmic}
\end{algorithm}

Different from previous works that only maximize the mutual information (MI) of the student output $r_{0}^{S} $ and the target $r_{n}^{T} $ from the EMA teacher, we propose to combine maximal the MI between $r_{0}^{S} $ and each $r_{j}^{T}, j \in [n-h,n-1] $. This is in the intuitive that we hope the student can synchronously learn instance temporal consistency from temporal knowledge. Our objective is as follows:

\begin{equation}\label{eq:tem}
\resizebox{0.9\linewidth}{!}{
$\mathcal{L}_{tem} \!= \mathop{max}\limits_{r_{0}^{S}}(I(r_{0}^{S};r_{n}^{T})\!+\! \sum_{j\!=\!n\!-\!h}^{n\!-\!1} I(r_{0}^{S};r_{j}^{T})) $
}
\end{equation}

The first term maximizes the MI in the current phase, like previous works do \cite{CPC1, CPC2, MOCO, BYOL}, which can only learn spatial consistent representations between different views. The second term maximizes the MI with previous knowledge to encourage the temporal consistency representations between different training stages, to mitigate the oscillation and catastrophic forgetting. Because the mutual information is notoriously hard to estimate, we instead maximizing the lower bound of it by the InfoNCE \cite{CPC1,MOCO, SimCLR}:
\begin{equation}
\label{eq:nce}
\resizebox{0.9\linewidth}{!}{
$\mathcal{L}_{tem}^{(1)} \!=\! \sum_{j\!=\!n\!-\!h}^{n} -log\frac{sim(r_{0}^{S} \cdot r_{j}^{T})}{sim(r_{0}^{S} \cdot r_{j}^{T}) \!+ \sum_{r_{j}^{-}} sim(r_{0}^{S} \cdot r_{ j}^{-})} $
}
\end{equation}
where $ r_{j}^{-} $ presents the representation of other samples from the same teacher $T_{j} $, and $sim(r_{0}^{S} \cdot r_{j}^{T})$ means their cosine similarity as following:
\begin{equation}
\label{eq:sim}
sim(r_{0}^{S} \cdot r_{j}^{T}) = exp(r_{0}^{S} \cdot r_{j}^{T}/\tau)
\end{equation}


\noindent where $\tau $ is temperature coefficient. In Eq. \ref{eq:nce}, the term  $j=n $ estimates the MI with the current target, the other terms estimate the MI with temporal targets. InfoNCE is relied on the negative samples to estimate the probability distributions. Furthermore, our methods can also work on the methods without negative samples like BYOL \cite{BYOL}. We minimize the L2 distance to maximize the MI for these works:
\begin{equation}\label{eq:l2}
\mathcal{L}_{tem}^{(2)}  = \sum_{j=n-h}^{n}|| r_{0}^{S}-r_{j}^{T}||^{2}
\end{equation}

\subsection{Overall Framework}
\label{method:Integral}

As previous works do, TKC also introduce a student $S(\cdot) $ amd an EMA teacher $T_{n}(\cdot) $. For a training sample $x $ from the data distribute, we obtain $r_{0}^{S} $ from $S $ and  $r_{n}^{T} $ from $T_{n} $.
To learn consistent knowledge, we also get targets from the \textit{temporal teachers} as $z_{j}^{T}, j \in [n-h, n-1] $. These targets should transport to the \textit{knowledge transformer} to filter important knowledge as $r_{j}^{T}, j \in [n-h, n-1].$ Then all the representations are fed into the \textit{temporal loss} in Eq. \ref{eq:tem}. During training, all the teachers are set stop-gradient. The loss will be back-propagated to update the student $S $ and the \textit{knowledge transformer}  $K_{j}, j \in [n-h, n-1] $. Algorithm \ref{algorithm} summarizes the algorithmic flow of the TKC procedure.



\section{Experiments}
In this section, we evaluate the quality of feature representation learned by our proposed TKC on several unsupervised benchmarks. We first follow standard linear evaluation protocol to assess the learned representations on ImageNet \cite{imagenet}. Then we transfer the pre-trained features to different downstream tasks, including object detection, instance segmentation, and semi-supervised classification. Finally, we perform a set of analysis studies to give an intuition of its performance. For brief-expression, all the experiments are based on MoCo v2 \cite{MOCO_v2} framework and ResNet-50 \cite{resnet} backbone unless otherwise stated.

\subsection{Evaluation on Linear Classification}
\label{exp:lc}
\begin{table}[htb]
\resizebox{\linewidth}{!}
{
\begin{tabular}{lcccc}
\hline
Method & architecture & epochs & Top-1 & Top-5 \\ \hline
Random & - & 200 & 5.6 & - \\
Supervised & - & 200 & 75.5 & - \\ \hline
\multicolumn{5}{l}{\textit{200 epoch training}} \\
LA \cite{LA} & R50 & 200 & 60.2 & - \\
CMC \cite{CMC} & R50(2x) & 200 & 64.4 & 88.2 \\
CPC v2 \cite{CPC2} & R50 & 200 & 63.8 & 85.3 \\
MoCo \cite{MOCO} & R50 & 200 & 60.6 & - \\
MoCHi \cite{MoCHi} & R50 & 200 & 68.0 &  \\
CO2 \cite{CO2} & R50 & 200 & 68.0 &  \\
MoCo v2 \cite{MOCO_v2} & R50 & 200 & 67.5 & - \\
\textbf{TKC} & R50 & 200 & \textbf{69.0}\textcolor[rgb]{0,0.7,0.3}{(+1.0)} & \textbf{88.7} \\ \hline
\multicolumn{5}{l}{\textit{400 epoch training}} \\
SwAV \cite{SWAV} & R50 & 400 & 70.1 & - \\
\textbf{TKC} & R50 & 400 & \textbf{70.8}\textcolor[rgb]{0,0.7,0.3}{(+0.7)} & \textbf{89.9} \\ \hline
\textit{800 $\& $ 1000 epoch training} &  &  &  &  \\
PIRL \cite{PIRL} & R50 & 800 & 63.6 & - \\
MoCo v2 \cite{MOCO_v2} & R50 & 800 & 71.1 & - \\
SimCLR \cite{SimCLR} & R50 & 1000 & 69.3 & 89.0 \\
\textbf{TKC} & R50 & 1000 & \textbf{72.1}\textcolor[rgb]{0,0.7,0.3}{(+1.0)} & 90.6 \\ \hline
\end{tabular}
}
\vspace{+1.2mm}
\caption{Top-1 and top-5 accuracy under the linear classification protocol on ImageNet with the MoCo framework and ResNet-50 backbone. We report our results of different epochs. }
\vspace{-3.5mm}
\label{table:resnet50}
\end{table}

We implement our TKC based on MoCo v2, which is composed of a standard ResNet-50 \cite{resnet} backbone and an MLP layer in the teacher-student framework. And the number of temporal teachers $h $ is set to 2. We train TKC model on 8 NVidia-1080ti GPUs with a mini-batch size of 256 and set $\alpha$ as 0.999, $\tau $ as 0.2. Moreover, we set the base learning rate $lr $ as 0.3, weight decay as 0.0001, and introduce a warm-up stage in the first 10 epochs, where linearly increase the learning rate from 0.01 to 0.03. All other hyper-parameters, training settings on pretext task and linear evaluation are strictly kept aligned with the implementations in \cite{MOCO_v2}. 

Table \ref{table:resnet50} summaries the top-1 and top-5 accuracy of our method. We report our results for different epochs pre-trained and also list top-performing methods. TKC improves MoCo v2 by 1.5 \% on 200 epochs results, which indicates that \textit{temporal teachers} can provide more accurate targets to learn consistent representations. Our results are also superiors to previous works on different pretext tasks, including all other instance discrimination paradigms. This demonstrates that temporal knowledge can benefit from stable training while mitigates the effect of catastrophic forgetting.

In order to verify the scalability of TKC, we respectively conduct our TKC on BYOL \cite{BYOL} baseline, and AlexNet \cite{alexnet} backbone. The number of teachers $h $ is changed to 3 for AlexNet. Specifically, We use a PyTorch implementation of BYOL in Momentum$_{2}$ Teacher \cite{M2T} as BYOL baseline and train the model for 100 epochs with 128 batch size on 8 Nvidia-1080ti GPUs. As for AlexNet, we adopt the implementation in Deep Clustering \cite{DC}, where we train the network with a mini-batch of 1024 on 4 NVidia-1080ti GPUs, and the learning rate is initialized by 0.24 with a cosine decay schedule for 200 epochs. More detail can be seen in the supplementary material.

\begin{table}[htb]
\resizebox{0.95\linewidth}{!}
{
\begin{tabular}{lcccc}
\hline
Method & architecture & epochs & Top-1 & Top-5 \\ \hline
BYOL$^{\dagger} $ \cite{BYOL}  & R50 & 100 & 70.1 & 90.6 \\
\textbf{BYOL$^{\dagger} $ + TKC} & R50 & 100 & \textbf{72.4}\textcolor[rgb]{0,0.7,0.3}{(+2.3)} & \textbf{91.7}\textcolor[rgb]{0,0.7,0.3}{(+1.1)} \\ \hline
\end{tabular}
}
\vspace{2mm}
\caption{Top-1 and top-5 accuracy under the linear classification protocol on ImageNet with BYOL framework. $^{\dagger} $ denotes the results from unofficial re-implementations.}
\label{table:byol}
\vspace{-3.5mm}
\end{table}

Table \ref{table:byol} shows our results on BYOL \cite{BYOL} baseline. We find that TKC can bootstrap BYOL for 2.3\%, which shows that temporal knowledge can also benefit different instance discrimination methods via maximizing the mutual information for temporal targets. The results of BYOL are incompatible with the official ones because we use an unofficial reproduction of BYOL. We conduct this experiment only to prove that TKC can improve different instance discrimination methods. For more details about this reproduction, please refer to the supplementary material.

\begin{table}[htb]
\resizebox{0.95\linewidth}{!}
{
\begin{tabular}{lccccc}
\hline
Method & conv1 & conv2 & conv3 & conv4 & conv5 \\ \hline
Random & 11.6 & 17.1 & 16.9 & 16.3 & 14.1 \\
Supervised & 19.3 & 36.3 & 44.2 & 48.3 & 50.5 \\ \hline
Jigsaw \cite{jigsaw} & 19.2& 30.1 & 34.7 & 33.9 & 28.3 \\
Rotation \cite{RotNet} & 18.8 & 31.7 & 38.7 & 38.2 & 36.5 \\
DeepCluster \cite{DC} & 12.9 & 29.2 & 38.2 & 39.8 & 36.1 \\
NPID \cite{NPID} & 16.8 & 26.5 & 31.8 & 34.1 & 35.6 \\
AET \cite{AET} & 19.2 & 32.8 & 40.6 & 39.7 & 37.7 \\
LA \cite{LA} & 14.9 & 30.1 & 35.7 & 39.4 & 40.2 \\
ODC \cite{ODC} & 19.6 & 32.8 & 40.4 & 41.4 & 37.3 \\
Rot-Decouple \cite{Rot3} & 19.3 & 33.3 & 40.8 & 41.8 & \bf{44.3} \\ \hline
\bf{TKC} & \bf{20.3}\textcolor[rgb]{0,0.7,0.3}{(+1.1)} & \bf{34.2}\textcolor[rgb]{0,0.7,0.3}{(+0.9)} & \bf{42.6}\textcolor[rgb]{0,0.7,0.3}{(+1.8)} & \bf{46.2}\textcolor[rgb]{0,0.7,0.3}{(+4.4)} & 44.0 \\ \hline
\end{tabular}
}
\vspace{2mm}
\caption{Top-1 accuracy under the linear classification protocol on ImageNet with the AlexNet backbone. We fine-tune a fc layer from the top of different layers.}
\vspace{-3.5mm}
\label{table:alexnet}
\end{table}

For AlexNet, as shown in Table \ref{table:alexnet}, TKC achieves state-of-the-art top-1 accuracy on conv1 to conv4, which outperforms all self-supervised methods on this track. Despite that TKC from conv5 underperforms Rot-decouple \cite{Rot3} by 0.3\%, our best result is from conv4, which surpasses the best of Rot-decouple by 1.9\%. The results show that TKC is also a leading method on AlexNet linear classification benchmark. We notice that TKC has more improvement on AlexNet than ResNet-50. This might be because the dropout layer in AlexNet can provide various temporal knowledge, which could be more effective in learning instance temporal consistency.

\subsection{Transfer to Downstream Tasks}

The primary goal of self-supervised learning is to learn good representations that transfer well on downstream tasks. In this subsection, we transfer the representations of 200 epoch TKC to three benchmarks: object detection, instance segmentation, and semi-supervised learning. We show that TKC learns better transferable representations on all three downstream tasks.

\noindent\textbf{Object Detection.} We both transfer to VOC \cite{voc} and COCO \cite{coco} dataset to evaluate our representations. As for Pascal VOC, We use Faster R-CNN \cite{faster} with ResNet50 backbone as the detector. We fine-tune the candidate pre-trained model for 48k iterations with a min-batch size of 8 on Pascal VOC \cite{voc} training set. The learning rate is initialized from 0.001 and then decayed at 36k and 44k iterations. The weight decay is set to 0.0001, and training image scales range between 480 to 800. We use $AP_{50}$, $AP $, $AP_{75}$ as evaluation metric on VOC test2007 set.

\begin{table}[bht]
\begin{center}
\resizebox{1.0\linewidth}{!}{
\begin{tabular}{ll|ccc}
\hline
& pre-train & $AP_{50}$ & $AP$ & $AP_{75}$ \\ \hline
& random-init & 60.2 & 33.8 & 33.1 \\
& supervised & 81.3 & 53.5 & 58.8 \\
& NPID++ \cite{NPID} & 79.1 & 52.3 & 56.9 \\
& PIRL \cite{PIRL} & 80.7 & 54.0 & 59.7 \\
& MoCo v2 \cite{MOCO} & 81.5 & 55.9 & 62.6 \\ \cline{2-5}
& TKC & \bf{81.8}\textcolor[rgb]{0,0.7,0.3}{(+0.3)} & \bf{56.5}\textcolor[rgb]{0,0.7,0.3}{(+0.6)} & \bf{62.8}\textcolor[rgb]{0,0.7,0.3}{(+0.2)} \\ \hline
\end{tabular}
}
\end{center}
\vspace{-2mm}
\caption{Object detection fine-tuned on PASCAL VOC with Faster-RCNN.}
\vspace{-3mm}
\label{table:voc-finetune}
\end{table}

\begin{table}[htb]
\begin{center}
\resizebox{1.0\linewidth}{!}{
\begin{tabular}{clccc}
\hline
& {pre-train} & $AP_{50}$ & $AP$ & $AP_{75}$ \\ \hline
& supervised & 59.8 & 40.2 & 43.8 \\
fine-tune & MoCo v2 & 60.0 & 40.1 & 43.4 \\ \cline{2-5}
& TKC & \bf{60.1}\textcolor[rgb]{0,0.7,0.3}{(+0.1)} & \bf{40.4}\textcolor[rgb]{0,0.7,0.3}{(+0.3)} & \bf{43.9}\textcolor[rgb]{0,0.7,0.3}{(+0.5)} \\ \hline
& supervised & 54.3 & 34.3 & 36.5 \\
freeze & MoCo v2 & 48.1 & 29.2 & 30.8 \\ \cline{2-5}
& TKC & \bf{54.2}\textcolor[rgb]{0,0.7,0.3}{(+6.1)} & \bf{34.7}\textcolor[rgb]{0,0.7,0.3}{(+5.5)} & \bf{37.1}\textcolor[rgb]{0,0.7,0.3}{(+6.3)} \\ \hline
\end{tabular}
}
\end{center}
\vspace{-2mm}
\caption{Object detection on COCO. The detection framework is Mask R-CNN. We report the results for both fine-tune and freezing the backbone.}
\vspace{-3mm}
\label{table:coco-detection}
\end{table}

For COCO \cite{coco} dataset, we train a Mask R-CNN \cite{mask} to learn the object detection and instance segmentation tasks synchronously. We train it for 180k iterations and decay the learning rate by 0.1 at 120k and 160k iterations. The input image size is between 640 and 800 on the training stage and 800 on the test stage. All hyper-parameters of this fine-tuning protocol are consistent with the MoCo v2 baseline.

As shown in Table \ref{table:voc-finetune}, our TKC achieves 81.8 $AP $ on the PASCAL VOC dataset, which outperforms all pre-trained models from competitors include the supervised ones. Our TKC shows consistent improvement for both $AP_{50}$, $AP $, $AP_{75}$, which shows that TKC indeed learns more consistent and transferable representations than MoCo v2. The upper part of the table \ref{table:coco-detection} shows the results on COCO, TKC as well surpass the MoCo v2 $AP_{75} $ by 0.5\% . The results on these two datasets indicated that comprehensive temporal knowledge can lead to transferable representations and learn better representations on different scenes and tasks.

\begin{table}[htb]
\begin{center}
\resizebox{1.0\linewidth}{!}{
\begin{tabular}{ll|ccc}
\hline
& pre-train & $AP_{50}$ & $AP$ & $AP_{75}$ \\ \hline
& random-init & 24.6 & 11.6 & 9.7 \\
& supervised & 80.2 & 51.4 & 55.5 \\
& MoCo v2 \cite{MOCO} & 79.0 & 51.7 & 56.2 \\ \cline{2-5}
& TKC & \bf{80.9}\textcolor[rgb]{0,0.7,0.3}{(+1.9)} & \bf{52.7}\textcolor[rgb]{0,0.7,0.3}{(+1.0)} & \bf{57.6}\textcolor[rgb]{0,0.7,0.3}{(+1.4)} \\ \hline
\end{tabular}
}
\end{center}
\vspace{-2mm}
\caption{Object detection on PASCAL VOC by freezing the backbone and only training the detection head of Faster-RCNN. }
\vspace{-3mm}
\label{table:voc-freeze}
\end{table}

We also evaluate TKC on detection in another way. We freeze the Faster R-CNN backbone and only train from the detection head to challenge it. This is somewhat like linear classification. Table \ref{table:voc-freeze} shows the results on VOC dataset. For both $AP $, $AP_{50} $ and $AP_{75} $, TKC surpass MoCo v2 baseline by more than 1.0\%, and also surpass the supervised counterpart. Table \ref{table:coco-detection} shows that on COCO dataset, the improvement is even more than 5.5 \%. Train on the frozen backbone can better reflect the pre-trained model's representation because that the trained head is more dependent on what the pretext task learns. The results on frozen backbone show that TKC does learn better semantics representations. The training detail can be seen in the supplementary material.

\noindent\textbf{Instance Segmentation.} We evaluate the instance segmentation on the COCO dataset, following the same setting as COCO detection. Table \ref{table:coco-segmentation} shows both the results by fine-tune and freezing. The finetune results gains 0.5\% $ AP_{75} $ on MoCo v2 baseline, indicates the temporal consistency can better locate the target to improve the IOU of instances. Moreover, the gain is further expanded to 1.9 \% when only train the segmentation head. We note in this way $AP_50 $ is improved by 3.0\%, show that TKC can also learn better representations on a simple task.

\begin{table}[htb]
\begin{center}
\resizebox{1.0\linewidth}{!}{
\begin{tabular}{clccc}
\hline
dataset & {pre-train} & $AP_{50}$ & $AP$ & $AP_{75}$ \\ \hline
& supervised & 56.7 & 34.9 & 37.1 \\
fine-tune & MoCo v2 & \bf{56.8} & 35.0 & 37.2 \\ \cline{2-5}
& TKC & \bf{56.8} & \bf{35.2}\textcolor[rgb]{0,0.7,0.3}{(+0.2)} & \bf{37.7}\textcolor[rgb]{0,0.7,0.3}{(+0.5)} \\ \hline
& supervised & 51.1 & 30.6 & 31.7 \\
freeze & MoCo v2 & 48.1 & 29.2 & 30.8 \\ \cline{2-5}
& TKC & \bf{51.1}\textcolor[rgb]{0,0.7,0.3}{(+3.0)} & \bf{30.9}\textcolor[rgb]{0,0.7,0.3}{(+1.7)} & \bf{32.7}\textcolor[rgb]{0,0.7,0.3}{(+1.9)} \\ \hline
\end{tabular}
}
\end{center}
\vspace{-2mm}
\caption{Instance segmentation on COCO. The detection framework is Mask R-CNN. We report the results for both fine-tune and freezing the backbone.}
\vspace{-3mm}
\label{table:coco-segmentation}
\end{table}

\begin{table}[htb]
\begin{center}
\resizebox{1.0\linewidth}{!}{
\begin{tabular}{lcccc}
\hline
\multirow{2}{*}{Method} & \multirow{2}{*}{Model} & \multirow{2}{*}{Epochs} & \multicolumn{2}{c}{Label fraction} \\
& & & 1\% & 10\% \\ \hline
Supervised & R50v2 & & 48.4 & 80.4 \\ \hline
NPID \cite{NPID} & R50 & 200 & 39.2 & 77.4 \\
PIRL \cite{PIRL} & R50 & 800 & 57.2 & 83.8 \\
MoCo v1\cite{MOCO}$^{\dagger}$ & R50 & 200 & 61.3 & 84.0 \\
SimCLR \cite{SimCLR}$^{\dagger}$ & R50 & 200 & 64.5 & 82.6 \\
MoCo v2 \cite{MOCO_v2}$^{\ddagger}$ & R50 & 200 & 61.7 & 84.6 \\ \hline
\textbf{TKC} & R50 & 200 & \textbf{72.1}\textcolor[rgb]{0,0.7,0.3}{(+10.4)} & \textbf{86.2}\textcolor[rgb]{0,0.7,0.3}{(+1.6)} \\ \hline 
\end{tabular}
}
\end{center}
\vspace{-1.2mm}
\caption{\textbf{Semi-supervised Learning on ImageNet.} We finetune the model with 1\% and 10\% labels. Center-crop top-5 accuracy is reported to compare with previous methods. $^{\dagger}$ indicates that the score is from this work \cite{InterCLR}. $^{\ddagger}$ means that we implement under the same strategy using the officially released pre-trained model.}
\label{table:semi}
\end{table}

\noindent\textbf{Semi-supervised Learning.} We then evaluate the utility of TKC in a data-efficient setting by performing semi-supervised learning on ImageNet. In this benchmark, We follow the experimental setup of \cite{SimCLR,PIRL}. The dataset is sampled of 1\% and 10\% from the labeled ImageNet-1k training data in a class-balanced way. We finetune the TKC pre-trained model on these two labeled subsets and validate it on the whole ImageNet validation data. In order to compare with previous works, we report the top-5 accuracy. The supervised baseline from \cite{s4l} is trained only using 1\% and 10\% labels, with a stronger architecture of ResNet50-v2, trained for 1000 epochs. Table \ref{table:semi} shows that our TKC surpasses all the previous methods trained for 200 epochs. When only 1\% of data is labeled, TKC surpasses our MoCo v2 baseline by a large margin of 9.6\%, indicating that the temporal knowledge is more beneficial when lacking labeled data. 

In addition, the mainstream semi-supervised learning methods adopt a consistent regularization to learn smooth manifold. The intuition in this field is similar to us, where they consider that consistent representation between similar samples can bring up accuracy classification boundary. Similarly, TKC also encourages consistent representations between different training stages to get a smoother manifold. The significant improvement on semi-supervised benchmarks shows we indeed learn temporal consistent representations.

\subsection{Analysis}

\noindent{\bf Ablation study.} Our method introduces two new hyper-parameters, the step interval of each teacher $s $, and amount of teachers $h $. We use $s $ as the steps among an epoch and do not tune it. For $h $, we take an ablation on the AlexNet backbone. In Table \ref{table:ablation_h}, the first column $h=1 $ means only the EMA teacher is used, which is an implementation of MoCo v2. We see that the \textit{temporal teacher} can boost accuracy from 39.9 to 42.2 when introduced only one temporal teacher. TKC achieves the best performance when maintaining three teachers, for this setting can acquire the most temporal knowledge to stable the target. When increasing $h $ even more, the result is declined unexpectedly. This might be because when involving the too old teachers, their representations are changed too much. It is hard to learn consistently with these teachers. Nonetheless, this confirms our motivation again that the inconsistency between different training stages has alleviated convergence.

\begin{table}[ht]
\begin{center}
\resizebox{0.8\linewidth}{!}
{
\begin{tabular}{c|ccccc}
\hline
$h $ & 1 & 2 & 3 & 4 & 6 \\ \hline
Top-1 & 39.9 & 42.2 & 43.5 & 41.9 & 41.8 \\ \hline
\end{tabular}
}
\end{center}
\caption{Ablation study on the effect of teacher numbers.}
\vspace{-3mm}
\label{table:ablation_h}
\end{table}

\begin{figure}[ht]
\resizebox{0.9\linewidth}{!}{
\includegraphics[]{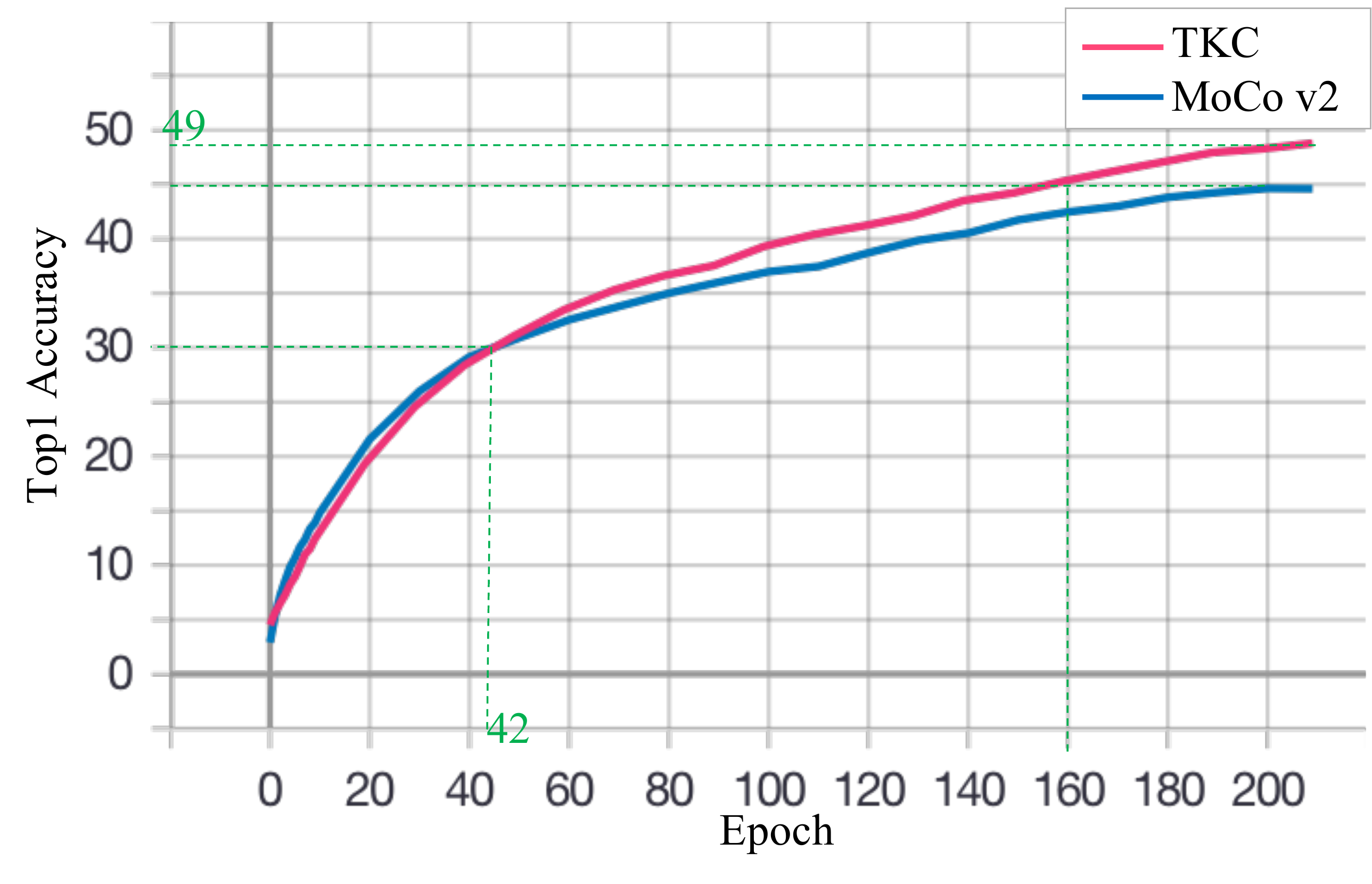}
}
\vspace{-2mm}
\caption{Comparison of validation accuracy between MoCo v2 and TKC. The top-1 accuracy is from a kNN classifier.}
\label{fig:ana2}
\end{figure}

\noindent{\bf Convergence comparison.} In Section \ref{method:loss}, We consider that TKC can combine maximize the mutual information with the target from the different stages, and therefore will enforce the network to learn temporal consistent representations and mitigate the catastrophic forgetting. To confirm our proposal, we use a $k$NN classifier to validate the model performance during training. As shown in Fig. \ref{fig:ana2}, TKC has a lower accuracy in the earlier training, which is because that the model is more inconsistent and noisy in the earlier stage, resulting in a big difference between the \textit{temporal teacher} and the current teacher. This difference prevents TKC from providing consistent signals. However, the TKC catches up MoCo v2 from the middle stage and finally surpasses it for 4.6\% at the end of the training, which indicates that TKC can stably provide a consistent signal from the middle training. This consistent signal can guide a more accurate training direct and accelerate convergence. Fig. \ref{fig:ana2} shows that the TKC at 160 epochs meets the accuracy of fully trained MoCo, reducing 80\% training time by mitigating catastrophic forgetting.

\section{Conclusion}

We summarize the existing instance discrimination methods into a teacher-student framework and note that the teacher can only provide instance spatial consistency. However, the output of the same instance can vary dramatically between different epochs when only spatial consistency is involved. We instead present a novel and strong method named Temporal Knowledge Consistency (TKC), which integrates the knowledge from previous teachers to improve the model's robustness and prevent possible catastrophic forgetting. TKC contains three modules. The \textit{temporal teacher} introduces the instance temporal consistency from previous models, the \textit{knowledge transformer} leverages the knowledge of these teachers, and the \textit{temporal loss} reduces the MI between the student and the temporal teacher. Temporal teacher is an orthogonal improvement for different instance discrimination methods. Our experimental results show that TKC can improve different frameworks MoCo, BYOL, and architectures ResNet-50, AlexNet. It also provides transferable representations on downstream tasks such as object detection, instance segmentation, and semi-supervised learning. Moreover, we hope our study can draw much attention to solve the unstable in unsupervised learning and search for effective ways to generate stable output with no labels.

{\small
\bibliographystyle{ieee_fullname}
\bibliography{egbib}
}

\clearpage

\appendix
\renewcommand{\appendixname}{Appendix~\Alph{section}}

\setcounter{figure}{3}  
\setcounter{table}{9}
\setcounter{equation}{7}  
\section{Implementation Details}

\subsection{Linear Classification.} 

For TKC on ResNet-50 \cite{resnet}, we freeze the ResNet-50 backbone and train a linear classifier after the frozen features from the global pooling layer. We use the student network as a pre-trained model. The classifier is trained for 100 epochs, with initialized learning rate $lr = 30$. We set momentum as 0.9, weight decay as 0, and decay the learning rate by 0.1 at the 60th epoch and 80th epoch. The batch size is set to 256 on 8 NVidia-1080ti GPUs.

\subsection{Based on BYOL}
\noindent\textbf{Pretext Training.} In the paper, we have implemented an experiment based on BYOL \cite{BYOL} to show that TKC can improve different methods (in Table \textcolor[RGB]{255,0,0}{2}). We use the BYOL baseline based on a pytorch implementation in Momentum$^{2} $ teacher \cite{M2T}. Their code is publicly available at \url{https://github.com/zengarden/momentum2-teacher}.

We use momentum SGD with momentum 0.9 and weight decay 1e-4. We train both the BYOL baseline and our TKC for 100 epochs, the basic learning rate is 0.05. We use a warm-up stage at the beginning of training for 10 epochs, and then cosine decays the learning rate. The batch size is 256 on 8 NVidia-1080ti GPUs. The data augmentation and the architecture are the same as the original paper \cite{BYOL}, except that we use batch normalization instead of SyncBN. The MLP projection head consists of two linear layers, with a batch norm layer and a ReLU layer between them. Our TKC+BYOL shares the same setting with the baseline, we set $h $ as 3, and also use a symmetrized loss.

\noindent\textbf{Linear Classification.} The setup of linear classification is also following the reproduction in \cite{M2T}. We fetch out the teacher encoder and freeze its backbone. Then we train a classifier consisting of a linear layer and a batch norm layer, following the global average pooling layer in the backbone. We train for 5 epochs. This reproduction is not strictly reimplemented the results in BYOL \cite{BYOL}, thus we do not compare this result with other methods. The results in Table \textcolor[RGB]{255,0,0}{2} show that TKC has good scalability, and can improve different methods.


\subsection{Based on AlexNet}

\noindent\textbf{Pretext Training.} We adopt the AlexNet \cite{alexnet} implementation in Deep Clustering \cite{DC} as the backbone, and additionally append a two-layers MLP behind it, following MoCo v2 \cite{MOCO_v2}. We train the model on 4 NVidia-1080ti GPUs with 1024 batch size. The learning rate in initialized as 0.24, and cosine decayed for 200 epochs. We set $\tau $ as 0.2, $\alpha $ as 0.9, following MoCo v2 \cite{MOCO_v2}. Differently, we set the number of negative samples $K $ as 8192 to accelerate training.

\noindent\textbf{Linear Classification.} In this section, we freeze the backbone of the student model and train a classifier containing a linear layer with 1000 output dimensions after the backbone for 100 epochs. The initial learning rate is set as $0.01 $, and decayed by 0.1 at the 60 $th $ epoch and 80 $th $ epoch.

\subsection{Linear Detection and Segmentation}

\noindent\textbf{Object Detection on PASCAL VOC.} We have performed linear detection on PASCAL VOC in Table \textcolor[RGB]{255,0,0}{5} to show that TKC learns better representations for object detection. We use Faster R-CNN \cite{faster} detector based on Detectron2 \cite{detectron}. The backbone ends with conv5 stage and is set frozen. The training hyperparameters are kept consistent with fine-tuning, except the backbone is frozen. We train it for 48k iterations on VOC07+12 trainval. The initial learning rate is 0.02 and decayed at 36k and 44k iterations. The warmup stage lasts for 200 iterations.

\noindent\textbf{Object Detection and Instance Segmentation on COCO.} We have also implemented linear detection on COCO. We use Mask R-CNN \cite{mask} detector based on Detectron2 \cite{detectron}, and synchronously train the object detection head and the instance segmentation head following \cite{MOCO}, meanwhile we freeze the backbone. We also use a 2x scheduler the same as finetuning, where we train it for 180k iterations. The size of the shorter side is in [640,800] pixels during training and is fixed as 800 at inference. We use 8 GPUs and 16 batchsize.

\section{Architecture of History Bank}

\begin{figure}[htb]
\centering
\resizebox{\linewidth}{!}
{
\includegraphics{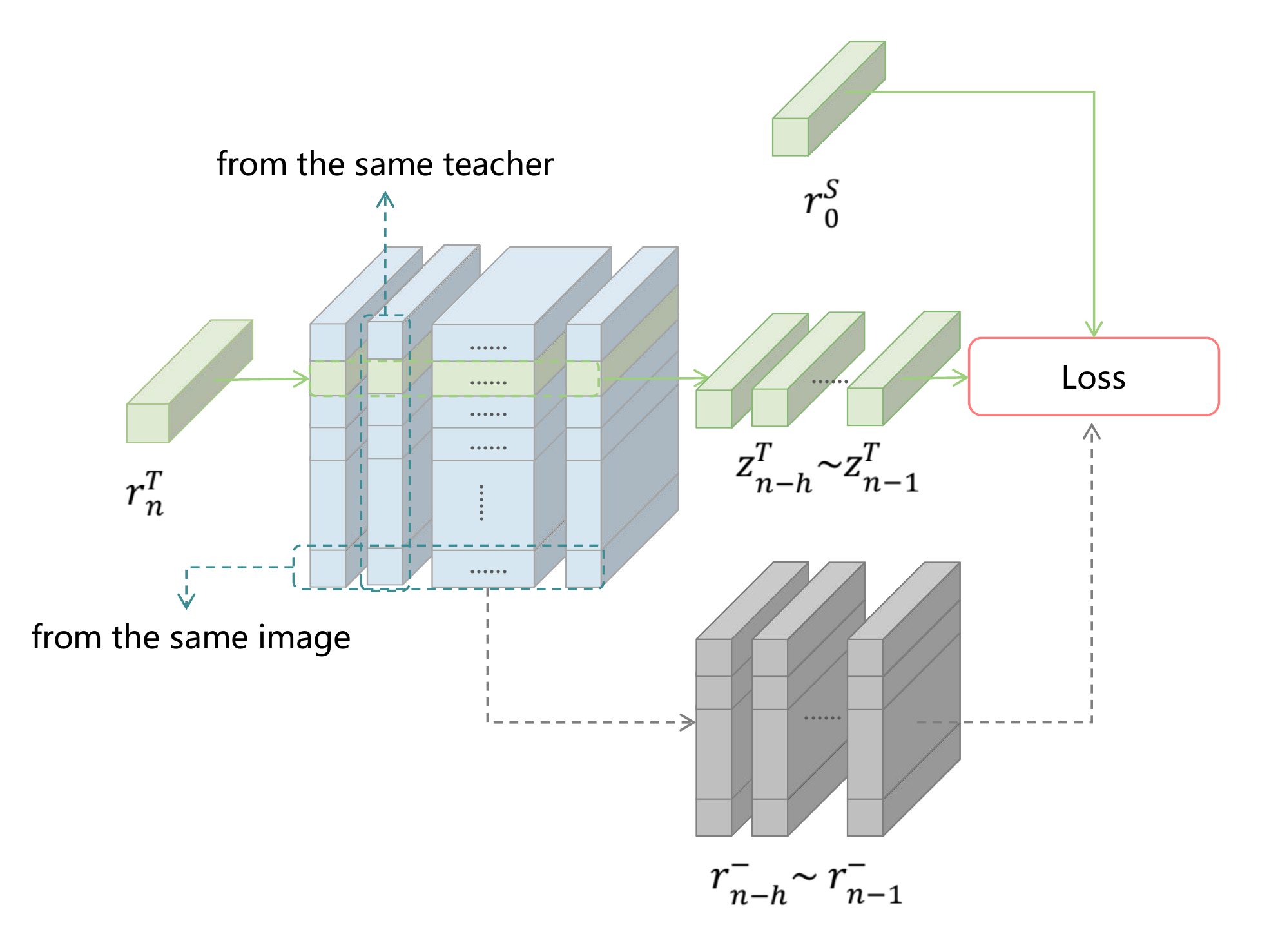}
}
\caption{\textbf{The architecture of history bank.} History bank is an effective implementation of the temporal teacher. A row in the history bank stores the features from the same image, a column in the history bank stores the features from the same teacher. The green cubes indicate positive features, the grey cubes indicate negative features. }
\label{fig:historybank}
\end{figure}

We use \textit{history bank} as a more effective implementation of the \textit{temporal teacher}. Fig \ref{fig:historybank} illustrates the architecture and mechanism of history bank. The history bank is a matrix with $size$ $ of $ $ \mathcal{D}$ rows and $h $ columns. $\mathcal{D} $ is the train set, $h $ is the number of \textit{temporal teachers}. A row of \textit{history bank} stores the features from the same image but different teachers, while a column of \textit{history bank } stores the ones from the same teacher but different images. This matrix can be saved at CPU memory, with no need to allocate the GPU memory. We illustrate the \textit{history bank} by the blue cube in Fig \ref{fig:historybank}.

For a sample $x $, we first get the feature $r_{0}^{S} $ from the student model, and the feature $r_{n}^{T} $ from the EMA teacher model. Then we fetch out all the features from the same image $x $ from the \textit{history bank} as $z_{n-h}^{T} \sim z_{n-1}^{T}$, as shown in the middle in the Fig \ref{fig:historybank}. For the implementation based on MoCo, we also fetch out the negative samples from the \textit{history bank}. For each $z_{j}^{T} j \in [n-h, n-1]$, the corresponding negative features are randomly selected from the same column in the \textit{history bank} as $r_{j}^{-}$. For the implementation based on BYOL, there is no need of the negative samples. \textit{History bank} is an effective implementation of \textit{temporal teacher}, the training procedure is the same as in the paper.

\section{Computational Cost}

We compare the computational cost of the two methods to show the efficiency of TKC. TKC use the \textit{history bank} to approximate the \textit{temporal teachers}. \textit{History bank} stores the features of the recent epochs in CPU memory, and only part of them corresponding to the current batch will be dumped to GPU memory. This optimization can avoid duplicate forwarding, and the features from \textit{history bank} are the same as forwarding the image into the \textit{temporal teachers}.

\begin{table}[htb]
\begin{center}
\resizebox{1.0\linewidth}{!}{
\begin{tabular}{llccc}
\hline
Method & \multicolumn{1}{c}{GPU} & batchsize & \multicolumn{1}{l}{GPU$\cdot$Time/Epoch} & \multicolumn{1}{l}{memory/GPU} \\ \hline
MoCo v2 & 8$\!\times\!$ 2080ti & 256 & 3.4h & 4.9G \\
\textbf{TKC} & 8$\!\times\!$ 2080ti & 256 & 5.0h & 5.0G \\ \hline
\end{tabular}
}
\end{center}
\caption{Computational cost. We report the time the GPU memory cost of our method and MoCo v2 baseline.}
\label{table:cost}
\end{table}

TKC has not too many additional costs thanks to \textit{history bank}. As shown in Table \ref{table:cost}, TKC has similar memory allocation with MoCo v2, which indicates that TKC has no special requirements for the capacity of machines. The time cost is higher than MoCo v2 for 47 \%, which is mainly from the matrix multiplications between temporal features $r_{j}^{T}, j \in [n-h, n-1]$ and $r_{0}^{S}$, the knowledge transformer, and the data transport between memory and GPU memory.

\section{Further analysis}
\subsection{More Experiment on AlexNet Backbone}

\begin{table}[bht]
\begin{center}
\resizebox{1.0\linewidth}{!}{
\begin{tabular}{lccccc}
\hline
Method & conv1 & conv2 & conv3 & conv4 & conv5 \\
\hline
MoCo v2 \cite{MOCO_v2} & 17.2 & 26.6 & 36.5 & 39.0 & \bf{42.8} \\
TKC & \bf{20.3}\textcolor[rgb]{0,0.7,0.3}{(+3.1)} & \bf{34.2}\textcolor[rgb]{0,0.7,0.3}{(+7.6)} & \bf{42.6}\textcolor[rgb]{0,0.7,0.3}{(+6.1)} & \bf{46.2}\textcolor[rgb]{0,0.7,0.3}{(+7.2)} & \bf{44.0}\textcolor[rgb]{0,0.7,0.3}{(+1.2)} \\ \hline
\end{tabular}
}
\end{center}
\caption{Comparison with MoCo v2 baseline on AlexNet.}
\label{table:compalex}
\end{table}

In this section, we implement MoCo v2 based on AlexNet \cite{alexnet} backbone to show that the temporal knowledge introduced by TKC can improve instance discrimination methods on the different backbone. In Table  \textcolor[RGB]{255,0,0}{3}, we only compare TKC with SOTA methods, here we supplement the result of MoCo v2. The MoCo v2 baseline follows the same setup with TKC. As shown in Table \ref{table:compalex}, TKC outperforms MoCo v2 baseline for all conv1 to conv5. The results from the bottom layers have more improvements, the results on conv4 especially surpass MoCo v2 for 7.2 \%.  This may because the temporal knowledge brought by the previous teachers can introduce the consistency between different epochs. The consistency can especially mitigate the 
dramatically changes and accelerates the convergence of the bottom layers.

\subsection{Relation to No EMA Methods}

Some recent works \cite{simsiam} claim that the EMA encoder is not necessary to prevent model collapse. However, their works have no conflict with our work. SimSiam has shown that the stop gradient but not the EMA encoder is the key to prevent model collapse, but it also admits that the EMA encoder can improve accuracy (in the last paragraph in Section 2). Table 4 in \cite{simsiam} reveal that SimSiam with EMA encoder (BYOL) surpasses it by 3.0\% for 800 epochs training, which shows that EMA encoder is important to learn good representations. However, the EMA encoder is not good enough, for it can not learn the temporal consistency between different training stages, as shown in our works.

We note that all the reproductions in \cite{simsiam} applies the symmetrized loss. Section 4.6 in \cite{simsiam} shows that the symmetrized loss can boost the accuracy for ~3\%, and the computational cost has also doubled. So comparing TKC which is asymmetric to the symmetric methods \cite{BYOL, simsiam} is unfair. Our improvement based on BYOL shows that the temporal knowledge is orthogonal to the symmetric loss. We consider providing the result of symmetric TKC in the next version and compare it with the symmetric methods.

\subsection{Does TKC improve the consistency? }
\begin{figure*}[htb]
\resizebox{\linewidth}{!}
{
\includegraphics{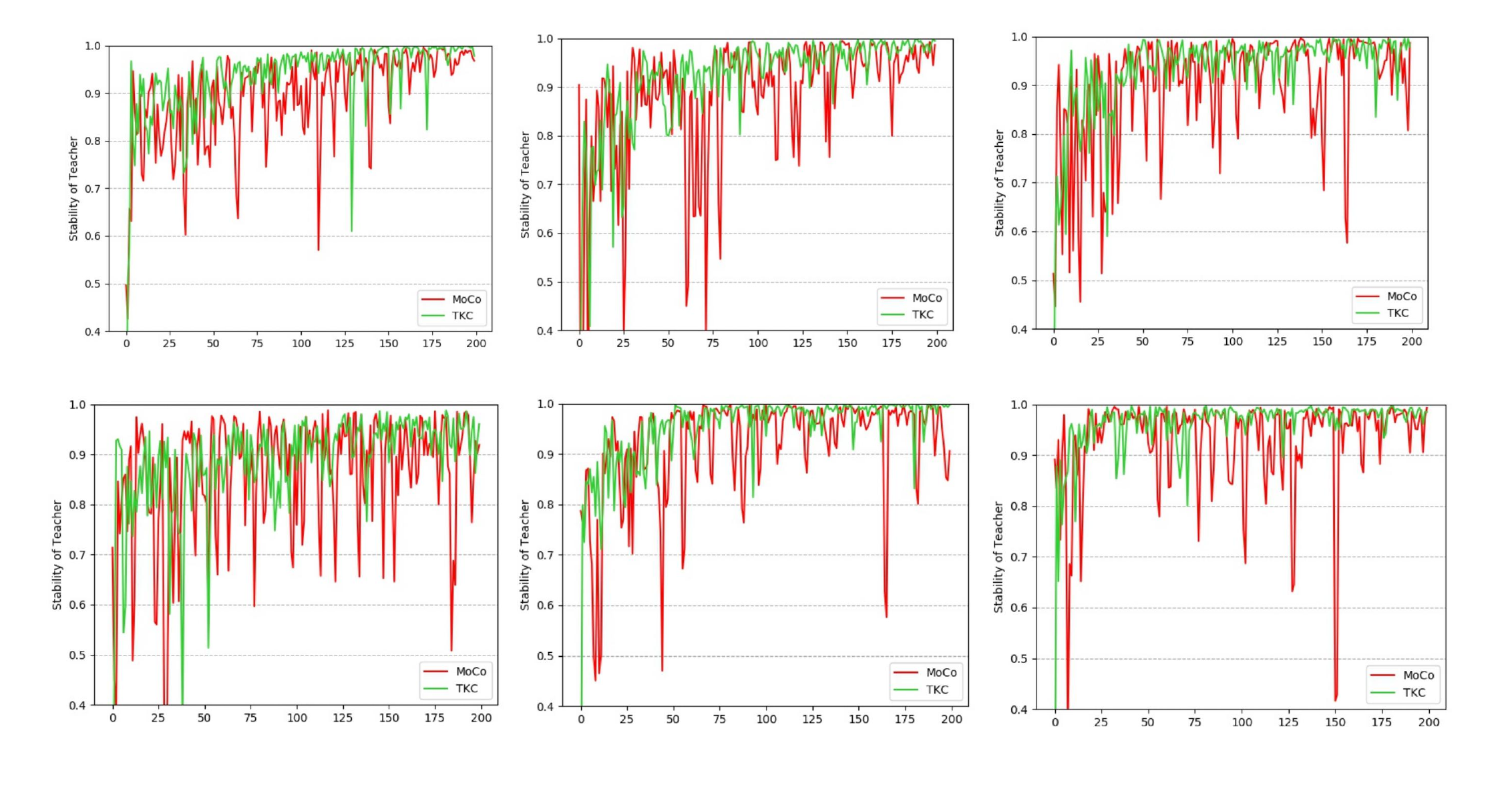}
}
\vspace{-5mm}
\caption{Comparison of stability between two methods in some randomly selected samples. The red curve represents MoCo v2, and the green curve represents TKC. The curves demonstrate that TKC can lead a consistent training and yield better representations.}
\vspace{-2mm}
\label{fig:stable}
\end{figure*}

In this section, we visualize the inconsistency during training and indicate that TKC can improve the stability during pretext training. We randomly select some images and compare the stability of each sample between TKC and MoCo v2 baseline.

Firstly, We define stability of a sample as the cosine similarity between current teacher output $r_{n}^{T} $ and the counterpart in the last epoch $z_{n-1}^{T} $. The formulation is:
\begin{equation}
stable(x) = r_{n}^{T} \cdot z_{n-1}^{T}
\label{eq:stable}
\end{equation}
Then we randomly select some samples from the training set and compute the stability of these samples respectively during the whole training procedure. We compute the stability for both MoCo v2 \cite{MOCO_v2} baseline and TKC. Each figure in Fig \ref{fig:stable} represents the stability of the same sample in different methods, the red curve represents MoCo v2, the green curve represents TKC.

As shown in Fig \ref{fig:stable}: (1) The output of the teacher model can dramatically vary even in the later stage during training. The stability of the samples usually gets down below 0.8 shows that the training target is inconsistent. Also, the stability is changed a lot in different epochs. These phenomena have confirmed our hypothesis that the targets from the teacher are noisy and inconsistent. (2) The stability of TKC is totally better than MoCo, where the green curve is higher than the red curve as a whole, which shows that the temporal knowledge from our method can lead to a more consistent training procedure, and improve the quality of the teacher's output.

\subsection{Ablation study about knowledge transformer}

\begin{table}[htb]
\center
\begin{tabular}{l|cc}
\hline
structure of KT & Top-1 & Top-5 \\ \hline
2-layer  & 65.91 & 87.07 \\
4-layer & 66.21 & 87.04 \\
2-layer bottleneck & 66.31 & 87.11 \\\hline
\end{tabular}
\vspace{+1.2mm}
\caption{Ablation study about knowledge transformer. All experiments are run on ResNet-50 for 100 epochs.}
\vspace{-3.5mm}
\label{table:ablation_kt}
\end{table}

In this section, we conduct ablation studies on different structures of the \textit{knowledge transformer}, as shown in the table \ref{table:ablation_kt}. All models are trained on ResNet50 for 100 epochs. In our work, we use an MLP to implement the knowledge transformer. This MLP consists of a linear layer with output dimension 256 followed by a ReLU nonlinearity and a final linear layer with output size 256. This structure can reach an accuracy of 65.91. We observe that increasing the layer of MLP can better extract the importance of different teachers to achieve better performance. The 4-layer MLP can further improve the top-1 accuracy to 66.31. And a design of bottleneck MLP can also boost the performance, where we change the hidden size from 256 to 4096 to obtain a bottleneck structure. This structure can boost the result to 66.21 with less additional computational cost. These results show that the \textit{temporal teacher} depends on the \textit{knowledge transformer} to leverage the importance of different teachers. Using more complex structures as attention may further improve the performance. We will explore it in future work.

\section{Difference with Related Works}
Some previous works also involve information from previous periods. MoCo \cite{MOCO} and Temporal Ensembling \cite{PI} both use the samples from previous training. In this section, we will clarify the difference between TKC and their works in both motivation and methodology.


\noindent\textbf{Difference with MoCo v2.} MoCo \cite{MOCO} believe that a large and consistent group of negative samples is critical for contrastive learning, and use the EMA encoder to construct a large and consistent negative bank. They think that the negative samples from previous training stages are harmful, and only use the negative samples which are near in time.

In our work, we notice that the outputs of the teacher can vary dramatically on the same sample during different training stages, which can introduce unexpected noise and lead to catastrophic forgetting caused by inconsistent objectives. We believe that the knowledge from previous stages is essential to learn the instance temporal consistency and stable the position of the teacher's outputs in the latent space. Empirically results show that the output of \textit{temporal teachers} can provide the temporal knowledge and gain the performance. Note that our negative bank is all consistent \cite{MOCO}, for we use the negative samples from the same teacher to compute the temporal loss in Eq \textcolor[RGB]{255,0,0}{4}.

\noindent\textbf{Difference with Temporal Ensembling.} Temporal Ensembling \cite{PI} is a semi-supervised learning method that ensemble the output of the same sample from previous epochs as the predicting target. Their work is different from ours in this aspect: (1) Temporal Ensembling relies heavily on dropout regularization to obtain various outputs in different epochs to yield a more accurate target. TKC also works well with networks without dropout layer \cite{resnet}, for TKC can restrict consistency between different epochs. (2) Temporal Ensembling also uses an exponential moving average to ensemble the output from different epochs, which can't leverage the importance of different outputs. On the contrary, TKC preserves the \textit{temporal teacher} independently and uses \textit{knowledge transformer} to dynamically learn their importance. 

\end{document}